\documentclass{article} 
\usepackage{nips12submit_e,times}
\usepackage{url}
\usepackage{listings, algorithm, algorithmic, graphicx, listing}

\title{Partitioning Large Scale Deep Belief Networks Using Dropout
}

\author{
Yanping Huang \\
University of Washington \\
huangyp@cs.washington.edu
\And
Sai Zhang \\
University of Washington\\
szhang@cs.washington.edu
}

\newcommand{\mnst}{{MNIST }}

\nipsfinalcopy 

\begin{document}
\maketitle

\begin{abstract}
Deep learning methods have shown great promise in many
practical applications, ranging from speech recognition,
visual object recognition, to text processing.
However, most of the current deep learning methods
suffer from scalability problems for large-scale applications,
forcing researchers or users to focus on small-scale
problems with fewer parameters.

In this paper, we consider a well-known  machine learning model, deep
belief networks (DBNs) that have  yielded impressive classification performance on a large number of benchmark machine learning tasks. To scale up DBN, we propose an approach that can use the computing clusters in a distributed environment to train large models, while the dense matrix computations within a single machine are sped up using graphics processors (GPU). When training a DBN, each machine randomly \textit{drops out} a portion of neurons in each hidden layer, for each training case, making the remaining neurons only learn to detect features that are generally helpful for producing the correct answer. Within our approach, we have developed four methods to combine outcomes from each machine to form a unified model.

Our preliminary experiment on the \mnst handwritten digit database demonstrates that our approach outperforms the state of the art test error rate.
\end{abstract}

\section{Introduction}
\label{sec:problem}

Deep learning methods~\cite{Hinton:2006:FLA} aim to learn a multilayer
neural network that can extract the feature hierarchies of the input data,  by maximizing the likelihood of its training data. Their promise largely lies in the potential to use vast amounts of unlabeled data to learn complex, highly nonlinear models with millions of parameters. In recent competitions, deep learning methods have shown adavantanges over nearest-neighbor based (shallow) methods such kernel methods~\cite{zhao2012fixed,scholkopf1998support} and ensemble methods~\cite{freund1997decision, Chen:ICML2013, Boosting15}. 

In this paper, we consider a well-known machine learning model, deep belief networks (DBNs), that can learn hierarchical representations of their inputs. DBN has been applied to a number of machine learning applications, including speech recognition~\cite{Deng_Speech12}, visual object recognition~\cite{abs-1003-0358, wu2014learning} and text processing~\cite{Bengio:2003:NPL}, among others. In particular, DBN is especially well-suited to problems with high-dimensional inputs, over which it can infer rich models with many hidden layers. For example, when
applied to images, a DBN can easily have tens of millions of free parameters, and ideally, we would want to use millions of unlabeled training examples to richly cover the input space.

It has been demonstrated that increasing the scale of
deep learning, with respect to the number of training examples,
the number of model parameters, or both, can drastically improve ultimate classification accuracy~\cite{ICML2011Le_210}. Unfortunately, with most of the current algorithms, even training a moderate-sized DBN can take weeks using a conventional implementation on a single CPU~\cite{Raina:2009:LDU}. This is primarily due to the daunting computational requirements in DBN training --- a large number of parameters need to be trained on the available examples.

To address the DBN scalability problem, this paper proposes an approach to scale up large-scale deep belief networks (DBNs) by adapting
the idea of \textit{random dropout}. \textit{Random dropout}, 
proposed by Hinton et al.~\cite{hinton:improving}, was originally used to prevent complex co-adaptations on the
training data in a single processor. On each training case, each hidden unit is randomly omitted
from the network with a probability of 0.5, so a hidden unit cannot rely on other hidden units
being present. By doing so, many separate DBNs are trained and then applied independently to the test data
to reduce the predication bias of a single DBN.

Our approach extends the \textit{random dropout} idea to the distributed and parallel setting.
Rather than omitting a hidden unit with a probability of 0.5, our approach \textit{randomly}
drops out \textit{a portion of} hidden units on each processor on each training case. To
combine DBNs in each processor, our approach offers four different ways (Section~\ref{sec:combination}):

\begin{enumerate}
\item performing model averaging with all trained DBNs.
\item using majority vote over the predication result of each trained DBN for each test case.
\item (each processor) \textit{synchronously} updating its parameters \textit{after} fetching the needed
parameters from other processors.
\item (each processor) \textit{asynchronously} fetching the computed parameters from other processors and pushing
its computed parameters to other processors.
\end{enumerate}

As validated in our preliminary evaluation, by using random dropout, our approach outperforms the state-of-the-art~\cite{hinton:improving} DBN algorithms on the same data set, and have the potential to exhibit nearly linear speedup with its parallel implementation.

This paper makes the following contributions:

\begin{itemize}
\item \textbf{Approach.} We propose an approach to scale up deep
belief networks using random dropout~\cite{hinton:improving} on large
clusters (Section~\ref{sec:approach}).

\item \textbf{Implementation.} We implemented our approach in
an open-source prototype, which is publicly available at: \url{http://deeplearning.googlecode.com}

\item \textbf{Evaluation.} We applied our approach to the
\mnst dataset, and demonstrated its effectiveness (Section~\ref{sec:evaluation}).
\end{itemize}

\section{Related Work}
\label{sec:related}

Recently, many approaches have been developed to scale up machine learning algorithms within a machine (via multithreading) and across machines (via message passing)~\cite{Langford09slowlearners, Mann09efficientlarge-scale, McDonald:2010:DTS, NIPS2010_1162}.    Much of the existing work focuses on linear, convex models, and takes distributed gradient computation as the first step. Some other approaches relax synchronization requirements, exploring delayed gradient updates for convex problems~\cite{Langford09slowlearners},  or exploring lock-less asynchronous stochastic gradient descent on shared-memory architectures (i.e. single machines)~\cite{Niu_NIPS11}.

Another way to scale up machine learning algorithms is to provide better abstractions and well-encapsulated computation tools. MapReduce~\cite{Dean:2008:MSD} and GraphLab~\cite{Gonzalez+al:osdi2012} are two notable examples.
However, MapReduce, originally designed for parallel data processing, has a number of limitations for training deep belief network~\cite{Gonzalez+al:osdi2012}. On the other hand, GraphLab~\cite{Gonzalez+al:osdi2012} was designed
for general \textit{unstructured} graph computations and does not exploit the computational effectiveness in a typical \textit{structured} graph as in a deep belief network. Thus, it is still unknown whether the abstraction of GraphLab can be used for training large-scale DBNs.

In the deep learning community, some work has been done to train relatively small models on a single machine~\cite{Hinton:2006:FLA}. In general, training a many-layer model is computationally intensive. Thus, full model parallelism as well as smart distributed optimization techniques is required. Recent years saw  a surge of interest in scaling up the training and inference algorithms used for DBNs~\cite{Dean:2008:MSD, Raina:2009:LDU}
and in improving applicable optimization procedures~\cite{ICML2011Le_210}. Existing approaches primarily
fall into the following two categories.

Approaches in the first category use graphics processors (GPUs)~\cite{Raina:2009:LDU, abs-1003-0358, Dahl12context} to achieve significant
speedup for training moderate-sized DBNs.  The use of GPUs has significantly reduced the computation time of matrix operations, which dominate most of the computation cost of deep learning algorithms.  However, a known limitation
of such GPU-based approaches is that the speedup will be small when the model does not fit in GPU memory (typically less than a few gigabytes). Thus, to effectively leverage a GPU, researchers often reduce the model size  and the parameter number to alleviate the impact of lacking enough GPU memory. While data and parameter reduction work well for small problems (e.g. acoustic
modeling for speech recognition~\cite{HintonDengSignal2012}), they are less attractive for realistic problems
with a large number of examples and dimensions (e.g., high-resolution images~\cite{Dean_NIPS12}).

Approaches in the second category use model parallelism to achieve scalability. For example, DistBelief~\cite{Dean_NIPS12} is a notable framework that enables model parallelism across machines (via message passing) , with the details of parallelism, synchronization and communication managed by the framework. Model parallelism under the DistBelief framework suffers a very large communication overhead due to the dense connections between layers of neurons. Data parallelism is also supported in DistBelief by using multiple replicas of a model to optimize a single objective. However, as pointed out by Hinton et al.~\cite{hinton:improving}, a large neural network, such as the one trained by DistBelief, can still perform poorly on held-out test data, if the relationship between the input and the correct output is complicated and the network has enough hidden units to model it accurately. In such cases, there will typically be many different
settings of the weights that can model the training set almost perfectly. Each of these weight vectors will make different predictions on held-out test data and
almost all of them will do worse on the test data than on
the training data because the feature detectors have been tuned to work well together on the training data but not on the test data.

Our approach is inspired by those above mentioned approaches,
and aims to address their limitations. With the goal of scaling up deep learning techniques to train very large DBNs, our approach combines the intrinsic parallelism in the ensemble learning algorithms, with  the random dropout approach~\cite{hinton:improving} to improve generalization results of neural networks. 
Using random dropout, our approach trains a separate DBN (much smaller than the original DBN) on an individual (graphical) processor  in a large cluster, and then combines their results using four proposed methods.  
Compared to existing approaches, our random dropout-based approach has several noticeable benefits. First, it becomes possible to train
a huge number of different networks in a reasonable time, since the number of parameters
to be trained on a single machine is much smaller than the original DBN.  Second, our approach permits to
better use the modern GPU memory due to the reduced model size.
Third, data transferring between processors would incur less communication overhead.

\section{Proposed Approach}
\label{sec:approach}

To train large DBNs, we propose an approach that 
supports distributed computation in neural networks.
At a high level, our approach consists of two steps: model parallelism (Section~\ref{sec:parallelism})
and model combination (Section~\ref{sec:combination}).
In the first model parallelism step, our approach
automatically parallelizes computation in each machine using all available resources,
and manages communication, synchronization and data transfer between machine.
In the second step, our approach supports four different ways to
combine results from each machine to form a unified DBN.

\subsection{Model Parallelism}
\label{sec:parallelism}

\begin{figure*}[!]
  \centering
  \includegraphics[scale=0.2000]{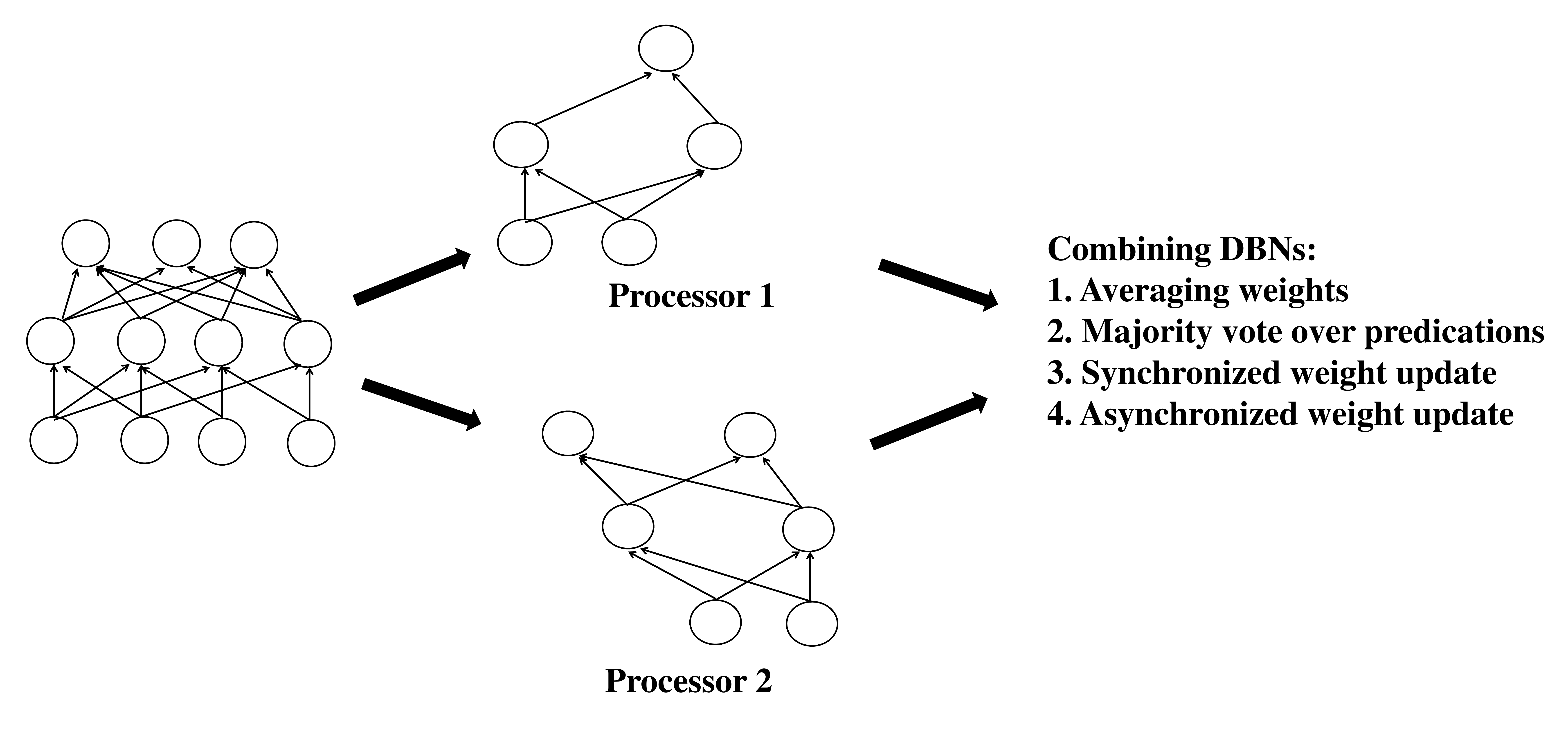}
  \vspace*{-2.0ex}\caption {{\label{fig:exampledbn} An example of model parallelism of a 3-layer neural network on 2 machines. In this example, each machine randomly drops out a portion of
  neurons and trains a separate DBN independently. Later, our approach combines
  the trained DBNs using four methods described in Section~\ref{sec:combination}.
}}
\end{figure*}

Figure~\ref{fig:exampledbn} illustrates the model parallelism step on two machines.

It is worthy noting that the computational needs of training a DBN on one machine
depends on its connectivity structure, and random dropout can significantly
reduce the complexity of a DBN as well as the number of parameters: dropping out
50\% of neurons at each layer can lead to a reduction of 75\% of the parameters (connection weights). In general, given a dropout probability $p$, our approach permits model parallelism among at most $1/p^2$  machines, with each machine updating a disjoint portion of weight matrix.
This is fundamentally different from existing model parallelism approaches~\cite{Dean_NIPS12}.
For example, in the DistBelief~\cite{Dean_NIPS12} framework, a user needs to define the computation that takes place at each machine in each layer of the model, while our approach distributes the computation of training each DBN fully automatically to each machine.
In addition, for a framework like DistBelief~\cite{Dean_NIPS12},
the complexity of a DBN is not reduced; rather, a DBN is partitioned (as a graph)
into available machines, and each machine must communicate frequently
(with the parameter server) for updating weights. Therefore,
large models with
 high computational demands might benefit from access to
more CPUs and memory at the beginning, but will be limited by
the bottleneck where communication costs dominate at some point.
By contrast, in our approach, DBNs produced by random dropout 
tend to be more amenable to extensive distribution than fully-connected structures, given
their less complex structures and lower communication requirements. Doing so
will also help alleviate the bottleneck in which many machines are waiting for the single
slowest machine to finish a given phase of computation.

\subsection{Model Combination}
\label{sec:combination}

Our approach provides four ways to combine the trained DBN
from each machine:

\begin{enumerate}
\item Averaging weights of the trained DBN in each machine.
\item Majority voting of the predication result for test data using the trained DBN in each machine
(our implementation breaks possible cycles by arbitrarily ranking the predication results, but this did not occur in our experiments).
\item \textit{Synchronously} updating parameter weights \textit{during} DBN training in each machine.
\item \textit{Asynchronously} updating parameter weights \textit{during} DBN training in each machine.
\end{enumerate}

The first two ways for model combination are straightforward, and are omitted in this paper
for space reasons. We next describe how to update parameter weights (a)synchronously.

\begin{figure}[t]
\begin{algorithmic}[1]
\STATE compute $\Delta$$w$
\COMMENT{calculate the gradient}
\STATE fetch $w$ from the weight queue
\STATE $w'$ $\leftarrow$ $w$ - $\eta$$\Delta$$w$
\STATE push $w'$ to other machines' weight queues
\vspace{-2mm}
\end{algorithmic}
\caption{The asynchronous parameter updating algorithm in each machine on each training data. Each
machine has a weight queue receiving the updated parameters from other machines.
At the beginning, each machine has the same replica of parameters.
} \label{fig:asynch}
\end{figure}

Figure~\ref{fig:asynch} sketches the asynchronous parameter weight updating algorithm in each
machine on each training data. This lock-free asynchronous algorithm prevents each machine from waiting for others to
finish before proceeding to the next training data, while sacrificing the data consistency of
parameters -- it is possible that two machines are updating the same parameters simultaneously
without an explicit ordering. As a consequence, this asynchronous algorithm may introduce additional
stochasticity in training.

\begin{figure}[t]
\begin{algorithmic}[1]
\STATE compute $\Delta$$w$
\COMMENT{calculate the gradient}
\STATE fetch $w$ from the parameter server
\COMMENT{wait until all other machines finish updating $w$}
\STATE $w'$ $\leftarrow$ $w$ - $\eta$$\Delta$$w$
\STATE update $w'$ on the parameter server
\COMMENT{wait until all other machines finish updating $w'$}
\vspace{-2mm}
\end{algorithmic}
\caption{The sychronous parameter updating algorithm in each machine on each training data. Only the parameter serve stores all weights; each machine fetches the needed weights from the server before updating.
} \label{fig:synch}
\end{figure}

Figure~\ref{fig:synch} sketches the synchronous parameter weight updating algorithm in each
machine. In this algorithm, there is a central parameter server storing
weights of all parameters. At the end of each mini-batch (a set of training
data), each machine sends a request to the parameter server to fetch the needed
parameter weights. If other machines are updating the requested parameters, this machine
needs to wait until all machines finish their updates. Comparing to the asynchronous algorithm,
this algorithm eliminates possible data races and improves data consistency
of the same parameters, but introduces higher overhead.


\subsection{Implementation}

We implemented our approach in a prototype using Matlab and Python.
Our prototype uses the Theano~\cite{theano} library to
define, optimize, and evaluate mathematical expressions involving multi-dimensional arrays
efficiently. Specifically, our implementation uses Theano to
achieve significant speedup in data-intensive calculations by leveraging the transparent use of a GPU.
When combining the trained DBNs, our implementation uses inter-process communication (IPC)~\cite{Stevens:2003} to exchange data among multiple threads
in one or more processes. Our implementation is publicly available at: \url{http://deeplearning.googlecode.com}.

\section{Evaluation}
\label{sec:evaluation}

\subsection{Details of  dropout training on MINST dataset}\label{sec:details}
The architectures of a deep network (the number of layers and the number of hidden units in each layer) varies on different benchmark tasks. Our first step is to develop a prototype and evaluate its effectiveness on the MNIST handwritten digits dataset~\cite{Hinton:2006:FLA}, which consists of 28 $\times$ 28 digit images, - 60,000 for training and 10,000 for testing.  The objective is to classify the digit images into their correct digit class. We experimented a neural network with size 784--500--500--2000--10, and pre-trained that network as a layer-wise Restricted Boltzmann Machine (RBM) for 50 epochs. Here one epoch  means a pass through training data. During the pre-training phase, we employs a greedy Contrastive Divergence--1 learning algorithm.  The learning rate is exponentially decaying, with initial value 10.0  and a decay rate 0.998 for each epoch of training. Weights were updated at the end of each mini-batch of size 100. Momentum is also used, with an initial value of 0.5, and a linear increasing rate of 0.001 over the first 490 epochs, after which it stays at 0.99.  For the fine tuning using back-propagation with dropout for 200 epochs, we  employed the stochastic gradient descent with mini-batches of size 100. We use the same dropout rates $p$ for all hidden units and $20\%$ dropout for input pixels. A constant learning rate 1.0 was used and there's no constraints imposed on the norms of weight vectors. 

\subsection{Generalization Performance as a function of dropout probability}
\begin{figure}
  \centering
   \includegraphics[scale=0.3]{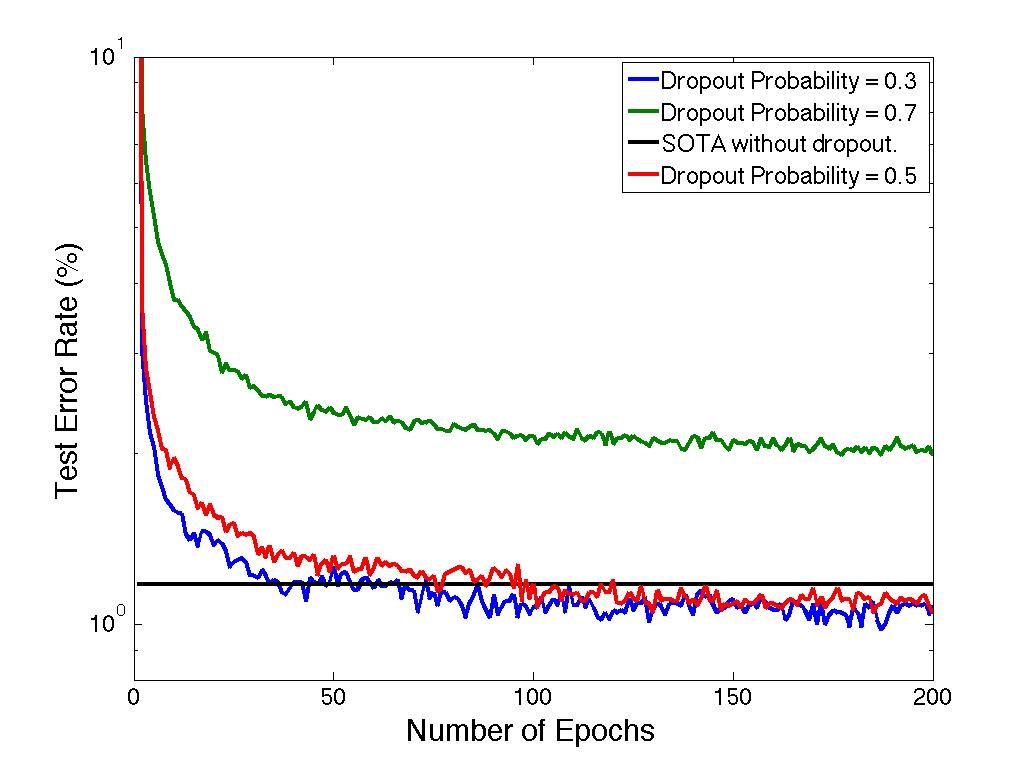}
  \label{fig:errorVSdropoutP}
   \caption{The test error rate on the \mnst for a variety of dropout probabilities. The input visible neurons also  use 20\% dropout. The best previous published results for this task using DBN without dropout are 1.18\% (with RBM pre-training) and 1.6\% (without RBM pre-training ), using more than 1,000 epochs~\cite{hinton:improving}.}
\end{figure}

In the original dropout~\cite{hinton:improving} article, Hinton et al claims that a better generalization performance can be achieved with various dropout probabilities. With implementation details shown in section~\ref{sec:details}, here we show how the test error rate varies as a function of dropout probability. As demonstrated in figure~\ref{fig:errorVSdropoutP}, dropout does decrease the test error rate by about 0.1\% (10 less misclassified examples in the test data set). Contrary to their claim,  we found that the generation performance of dropout actually depends on the dropout probability. When the dropout probability is greater than 0.6, the test error rate increases significantly. Such inconsistency might be due to the much smaller training epochs (1,000 epochs used in Hinton's paper) used in our implementation. 

\subsection{Generalization Performance  of our  distributed algorithm}
Here we evaluate the generalization performance on the \mnst dataset using various algorithms to combine results from different machines, as listed in section~\ref{sec:combination},.

\begin{table}[h!]
  \centering
  \begin{tabular}[h!]{|c|c|c|c|c|c|}\hline
   Algorithms & Sequential & Weight Averaging & Majority Vote & Synchronous Update & Asynchronous Update \\ \hline
 Error Rate (\%) & 1.08 & 0.98 & 1.04 & 0.97 & 1.06 \\ \hline
  \end{tabular}
  \caption{Test error rate using different algorithms for 200 epochs of fine tuning. Weight averaging and majority vote algorithms collect final weights from $7$ independent runs of the standard dropout algorithms. Synchronous update and asynchronous update algorithms combine results from two processes after each input instance. Dropout rate is 50\% for all algorithms. }
\end{table}

\begin{figure}
  \centering
     \includegraphics[scale=0.3]{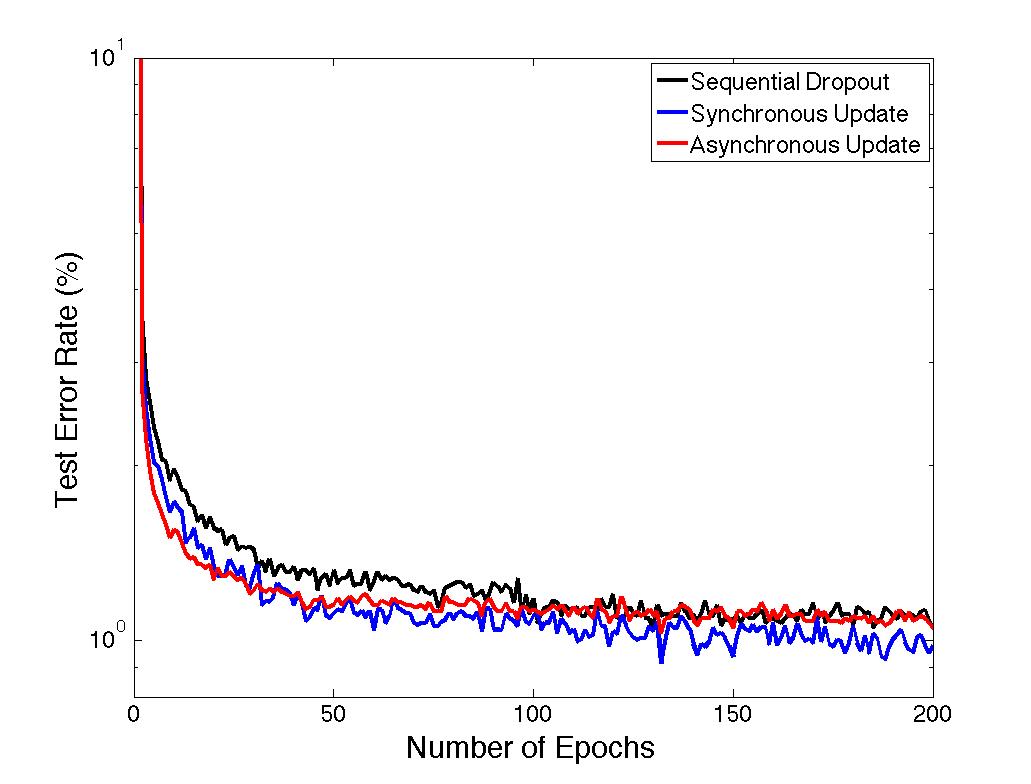}
  \caption{The test error rate on the \mnst as a function of number of epochs, using synchronous/asynchronous updates. }
  \label{fig:asyncUpdateVsTime}
\end{figure}

Both weight averaging and synchronous update algorithms achieved notable improvement in the generalization performance. Surprisingly the majority vote method didn't reduce the test error rate by a large margin. The asynchronous update algorithm introduced additional noise in the weight updates by its lock free mechanism, thus its generalization performance was relatively the same as the sequential dropout algorithm. However, as exhibited in  figure~\ref{fig:asyncUpdateVsTime}, the convergence rate for both synchronous and asynchronous update algorithms are faster than the 
sequential dropout algorithm.

\subsection{Limitations}

Our current evaluation was performed on a  desktop with a Dual Core Intel E7400 processor, 3GB RAM, and a NVIDIA 8800GS graphics card.  Pretraining/fine tuning are generally very time consumption on this  machine.  Due to time constraints, we were only able to evaluate our proposed algorithm on the relatively small \mnst dataset. However, we plan to further evaluate our algorithm on other speech and object recognition benchmark tasks such as the TIMIT Acoustic-Phonetic Continuous Speech Corpus~\cite{Mohamed12}, Reuters Corpus for news article topic recognition~\cite{Rose02thereuters}, and the ImageNet dataset of millions of labeled images in thousands of categories~\cite{Deng09imagenet:a}.

\section{Conclusion}
\label{sec:conclusion}

This paper proposes an approach to scale up deep belief networks (DBNs).
At the core of our approach is the use of \textit{random dropout} to
prevent co-adaptions on the training data for a DBN, reduce overfitting,
and enable DBN training to use the computational power of clusters in
a distributed environment.  Empirically, our implementation
outperforms some state-of-the-art approaches~\cite{hinton:improving}, and promising nearly linear speedups. Furthermore, our approach allows parallel training of a DBN even when some gradients are
computationally intensive.

For future work, it would be interesting to compare our approach
with other approaches using different abstractions~\cite{Gonzalez+al:osdi2012, graphchi:osdi2012}.
For example, the PowerGraph~\cite{Gonzalez+al:osdi2012} abstraction exploits the internal
structure of graph programs to address the challenges of
computation on natural graphs. Thus, it may be possible to
adapt a similar random dropout idea to reduce memory consumption
and communications between processors. An investigation
into how to generalize this approach to
other structures and problems would enable even faster computation of machine learning problems.

\bibliographystyle{unsrt} 
\bibliography{ref}

\begin{thebibliography}{10}

\bibitem{Hinton:2006:FLA}
Geoffrey~E. Hinton, Simon Osindero, and Yee-Whye Teh.
\newblock A fast learning algorithm for deep belief nets.
\newblock {\em Neural Comput.}, 18(7):1527--1554, July 2006.

\bibitem{zhao2012fixed}
Songlin Zhao.
\newblock {\em From fixed to adaptive budget robust kernel adaptive filtering}.
\newblock University of Florida, 2012.

\bibitem{scholkopf1998support}
Bernhard Sch{\"o}lkopf and Alex Smola.
\newblock Support vector machines.
\newblock {\em Encyclopedia of Biostatistics}, 1998.

\bibitem{freund1997decision}
Yoav Freund and Robert~E Schapire.
\newblock A decision-theoretic generalization of on-line learning and an
  application to boosting.
\newblock {\em Journal of computer and system sciences}, 55(1):119--139, 1997.

\bibitem{Chen:ICML2013}
Tianqi Chen, Hang Li, Qiang Yang, and Yong Yu.
\newblock General functional matrix factorization using gradient boosting.
\newblock In {\em Proceeding of 30th International Conference on Machine
  Learning (ICML'13)}, volume~1, pages 436--444, 2013.

\bibitem{Boosting15}
Nan Wang.
\newblock A new boosting algorithm based on dual averaging scheme.
\newblock {\em International Journal of Innovative Science and Modern
  Engineering}, 3(9):18--22, 2015.

\bibitem{Deng_Speech12}
Li~Deng, Dong Yu, and J.~Platt.
\newblock Scalable stacking and learning for building deep architectures.
\newblock In {\em Acoustics, Speech and Signal Processing (ICASSP), 2012 IEEE
  International Conference on}, pages 2133 --2136, march 2012.

\bibitem{abs-1003-0358}
Dan~Claudiu Ciresan, Ueli Meier, Luca~Maria Gambardella, and J{\"u}rgen
  Schmidhuber.
\newblock Deep big simple neural nets excel on handwritten digit recognition.
\newblock {\em CoRR}, abs/1003.0358, 2010.

\bibitem{wu2014learning}
Jia Wu, Raymond Tse, and Linda~G Shapiro.
\newblock Learning to rank the severity of unrepaired cleft lip nasal deformity
  on 3d mesh data.
\newblock In {\em Pattern Recognition (ICPR), 2014 22nd International
  Conference on}, pages 460--464. IEEE, 2014.

\bibitem{Bengio:2003:NPL}
Yoshua Bengio, R{\'e}jean Ducharme, Pascal Vincent, and Christian Janvin.
\newblock A neural probabilistic language model.
\newblock {\em J. Mach. Learn. Res.}, 3:1137--1155, March 2003.

\bibitem{ICML2011Le_210}
Quoc Le, Jiquan Ngiam, Adam Coates, Ahbik Lahiri, Bobby Prochnow, and Andrew
  Ng.
\newblock On optimization methods for deep learning.
\newblock In Lise Getoor and Tobias Scheffer, editors, {\em Proceedings of the
  28th International Conference on Machine Learning (ICML-11)}, ICML '11, pages
  265--272, New York, NY, USA, June 2011. ACM.

\bibitem{Raina:2009:LDU}
Rajat Raina, Anand Madhavan, and Andrew~Y. Ng.
\newblock Large-scale deep unsupervised learning using graphics processors.
\newblock In {\em Proceedings of the 26th Annual International Conference on
  Machine Learning}, ICML '09, pages 873--880, New York, NY, USA, 2009. ACM.

\bibitem{hinton:improving}
Geoffrey~E. Hinton, Nitish Srivastava, Alex Krizhevsky, Ilya Sutskever, and
  Ruslan Salakhutdinov.
\newblock Improving neural networks by preventing co-adaptation of feature
  detectors.
\newblock {\em CoRR}, pages --1--1, 2012.

\bibitem{Langford09slowlearners}
John Langford, Er~J. Smola, and Martin Zinkevich.
\newblock Slow learners are fast.
\newblock In {\em In NIPS}, pages 2331--2339, 2009.

\bibitem{Mann09efficientlarge-scale}
Gideon Mann, Ryan Mcdonald, Mehryar Mohri, Nathan Silberman, and Daniel~D.
  Walker.
\newblock Efficient large-scale distributed training of conditional maximum
  entropy models.
\newblock In {\em In Advances in Neural Information Processing Systems}, 2009.

\bibitem{McDonald:2010:DTS}
Ryan McDonald, Keith Hall, and Gideon Mann.
\newblock Distributed training strategies for the structured perceptron.
\newblock In {\em Human Language Technologies: The 2010 Annual Conference of
  the North American Chapter of the Association for Computational Linguistics},
  HLT '10, pages 456--464, Stroudsburg, PA, USA, 2010. Association for
  Computational Linguistics.

\bibitem{NIPS2010_1162}
Martin Zinkevich, Markus Weimer, Alex Smola, and Lihong Li.
\newblock Parallelized stochastic gradient descent.
\newblock In J.~Lafferty, C.~K.~I. Williams, J.~Shawe-Taylor, R.S. Zemel, and
  A.~Culotta, editors, {\em Advances in Neural Information Processing Systems
  23}, pages 2595--2603, 2010.

\bibitem{Niu_NIPS11}
Feng Niu, Benjamin Recht, Christopher Re, and Stephen~J. Wright.
\newblock Hogwild!: A lock-free approach to parallelizing stochastic gradient
  descent.
\newblock In {\em In NIPS}, pages 2331--2339, 2011.

\bibitem{Dean:2008:MSD}
Jeffrey Dean and Sanjay Ghemawat.
\newblock Mapreduce: simplified data processing on large clusters.
\newblock {\em Commun. ACM}, 51(1):107--113, January 2008.

\bibitem{Gonzalez+al:osdi2012}
Joseph~E. Gonzalez, Yucheng Low, Haijie Gu, Danny Bickson, and Carlos Guestrin.
\newblock Powergraph: Distributed graph-parallel computation on natural graphs.
\newblock In {\em Proceedings of the 10th {USENIX} {S}ymposium on {O}perating
  {S}ystems {D}esign and {I}mplementation ({OSDI} '12)}, Hollywood, October
  2012.

\bibitem{Dahl12context}
George~E. Dahl, Student Member, Dong Yu, Senior Member, Li~Deng, and Alex
  Acero.
\newblock Context-dependent pre-trained deep neural networks for large
  vocabulary speech recognition.
\newblock In {\em IEEE Transactions on Audio, Speech, and Language Processing},
  2012.

\bibitem{HintonDengSignal2012}
L.~Deng, D.~Yu, G.~Dahl, A.~Mohamed, N.~Jaitly, A.~Senior, V.~Vanhoucke,
  P.~Nguyen, T.~Sainath, and B.~Kingsbury.
\newblock Deep neural networks for acoustic modeling in speech recognition.
\newblock {\em IEEE Signal Processing Magazine}, 2012.

\bibitem{Dean_NIPS12}
Jeffrey Dean, Greg Corrado, Rajat Monga, Kai Chen, Matthieu Devin, Quoc Le,
  Mark Mao, Marc'Aurelio Ranzato, Andrew Senior, Paul Tucker, Ke~Yang, and
  Andrew Ng.
\newblock Large scale distributed deep networks.
\newblock In {\em Advances in Neural Information Processing Systems 25}, 2012.

\bibitem{theano}
The {T}heano {L}ibrary.
\newblock \url{http://deeplearning.net/software/theano/}.

\bibitem{Stevens:2003}
W.~Richard Stevens, Bill Fenner, and Andrew~M. Rudoff.
\newblock {\em UNIX Network Programming, Vol. 1}.
\newblock Pearson Education, 3 edition, 2003.

\bibitem{Mohamed12}
A.~Mohamed, G.~Dahl, and G.~Hinton.
\newblock Acoustic modeling using deep belief networks.
\newblock {\em IEEE Transactions on Audio, Speech, and Language Processing},
  20(14), 2012.

\bibitem{Rose02thereuters}
Tony Rose, Mark Stevenson, and Miles Whitehead.
\newblock The reuters corpus volume 1 - from yesterdayÕs news to tomorrowÕs
  language resources.
\newblock In {\em In Proceedings of the Third International Conference on
  Language Resources and Evaluation}, pages 29--31, 2002.

\bibitem{Deng09imagenet:a}
Jia Deng, Wei Dong, Richard Socher, Li~jia Li, Kai Li, and Li~Fei-fei.
\newblock Imagenet: A large-scale hierarchical image database.
\newblock In {\em In CVPR}, 2009.

\bibitem{graphchi:osdi2012}
Aapo Kyrola, Guy Blelloch, and Carlos Guestrin.
\newblock Graphchi: Large-scale graph computation on just a pc.
\newblock In {\em Proceedings of the 10th {USENIX} {S}ymposium on {O}perating
  {S}ystems {D}esign and {I}mplementation ({OSDI} '12)}, Hollywood, October
  2012.

\end{thebibliography}
\end{document}